\documentclass[letterpaper, 10 pt, conference]{ieeeconf}  

\IEEEoverridecommandlockouts                              

\newcommand\blfootnote[1]{
    \begingroup
    \renewcommand\thefootnote{}\footnote{#1}
    \addtocounter{footnote}{-1}
    \endgroup
}
\usepackage{url}
\usepackage{multirow}
\usepackage{graphicx}
\usepackage{algorithm}
\usepackage[noend]{algpseudocode}
\usepackage{subcaption}
\usepackage{amsmath}
\usepackage{amssymb}
\usepackage[table]{xcolor}
\usepackage{pgfplots}
\usepackage{tikz}
\pgfplotsset{compat=1.16}
\usetikzlibrary{positioning,patterns,patterns.meta,calc,arrows.meta}
\let\labelindent\relax
\usepackage{enumitem}
\usepackage[normalem]{ulem}  
\usepackage{wrapfig}
\usepackage{balance}

\newcommand{\OOD}{OOD }

\DeclareMathOperator*{\argmax}{arg\,max}
\DeclareMathOperator*{\argmin}{arg\,min}

\title{\LARGE \bf
A Metacognitive Approach to\\Out-of-Distribution Detection for Segmentation
}

\author{
  Meghna Gummadi, Cassandra Kent, Karl Schmeckpeper, and Eric Eaton\\
  University of Pennsylvania, Philadelphia, PA, USA \\
  \texttt{\{meghnag, dekent, karls, eeaton\}@seas.upenn.edu} \\
\thanks{*This research was partially supported by DARPA SAIL-ON contract HR001120C0040 and the Army Research Office MURI W911NF20-1-0080.}
}
\begin{document}

\maketitle
\thispagestyle{empty}
\pagestyle{empty}

\begin{abstract}

Despite outstanding semantic scene segmentation in closed-worlds, deep neural networks segment novel instances poorly, which is  required for autonomous agents acting in an open world. To improve out-of-distribution (OOD) detection for segmentation, we introduce a metacognitive approach in the form of a lightweight module that leverages entropy measures, segmentation predictions, and spatial context to characterize the segmentation model's uncertainty and detect pixel-wise OOD data in real-time. Additionally, our approach incorporates a novel method of generating synthetic OOD data in context with in-distribution data, which we use to fine-tune existing segmentation models with maximum entropy training. This further improves the metacognitive module's performance without requiring access to OOD data while enabling compatibility with established pre-trained models. Our resulting approach can reliably detect OOD instances in a scene, as shown by state-of-the-art performance on OOD detection for semantic segmentation benchmarks.

\end{abstract}


\section{INTRODUCTION}
\blfootnote{\textbf{\\This work has been submitted to the IEEE for possible publication. Copyright may be transferred without notice, after which this version may no longer be accessible.}}
Current deep neural networks (DNNs) achieve near-perfect performance in semantic segmentation, but only in the closed-world paradigm, where test data is drawn from the same distribution as training data \cite{zhu2019improving, Cheng_2020_CVPR, li2020improving, wang2020deep}. However, most real-world agents must operate in highly dynamic open-world settings, including autonomous driving~\cite{wong2020identifying, balasubramanian2021open}, assistive robotics~\cite{boteanu2015towards, feng2019challenges, kim2022learning}, and social robotics~\cite{gunther2017toward, dey2023addressing}, where encountering objects beyond the fixed training distribution is the norm. These deployed agents must behave reasonably on out-of-distribution (OOD) instances encountered in the open world. Training agents on large amounts of annotated data fails to generalize to unseen classes \cite{6365193, Bendale_2016_CVPR, Bendale_2015_CVPR}, but a practical alternative is to 
recognize novel or OOD instances as they are encountered and then accommodate the new data \cite{ganapini2023thinking, gummadi2022shels}. 
We focus on the first step: enabling semantic segmentation to identify OOD instances. 


An intuitive way of tackling this challenge is to look for low-confidence segmentations, 
based on the assumption that predictions will be less certain for OOD instances \cite{hendrycks2016baseline, liang2017enhancing}.
Reasoning over uncertainty metrics, including 
prediction dispersion \cite{chan2021entropy} and model consensus \cite{kendall2015bayesian,dietterich2000ensemble}, 
endows 
{\em metacognitive} systems~\cite{JOHNSON2022105743, schmill201112} to monitor their own performance and identify errors. A common approach is to make decisions using a simple threshold over the uncertainty quantification; while often effective, simple thresholding is non-adaptive and may overly generalize. We propose to provide segmentation models with a lightweight decision making module that considers context along with proven uncertainty measures, enabling segmentation models to perform 
metacognitive reasoning about their predictions.

Such metacognitive reasoning heavily relies on uncertainty estimates, however large DNNs are known to be overconfident, which may hinder detection of OOD instances~\cite{guo2017calibration}. Network calibration using an appropriate validation set~\cite{guo2017calibration} is a widely accepted way to reduce this overconfidence, but this raises the question: what validation data is appropriate to use? Top-performing OOD detectors assume access to sample OOD validation sets for network calibration~\cite{chan2021entropy}, hyperparameter tuning~\cite{liang2017enhancing}, and additional training~\cite{hendrycks2018deep}.
While these methods show that using OOD data during training is effective, the choice of such data can bias a model toward whatever a data engineer considers most likely to be encountered. 
Additionally, actual OOD data is not typically available at design time, and 
can not fully represent OOD data encountered after deployment. 

\begin{figure}
        \centering
        \hspace{-.65cm}
        \begin{subfigure}[b]{0.22\textwidth}
            \centering
            \includegraphics[width=\textwidth]{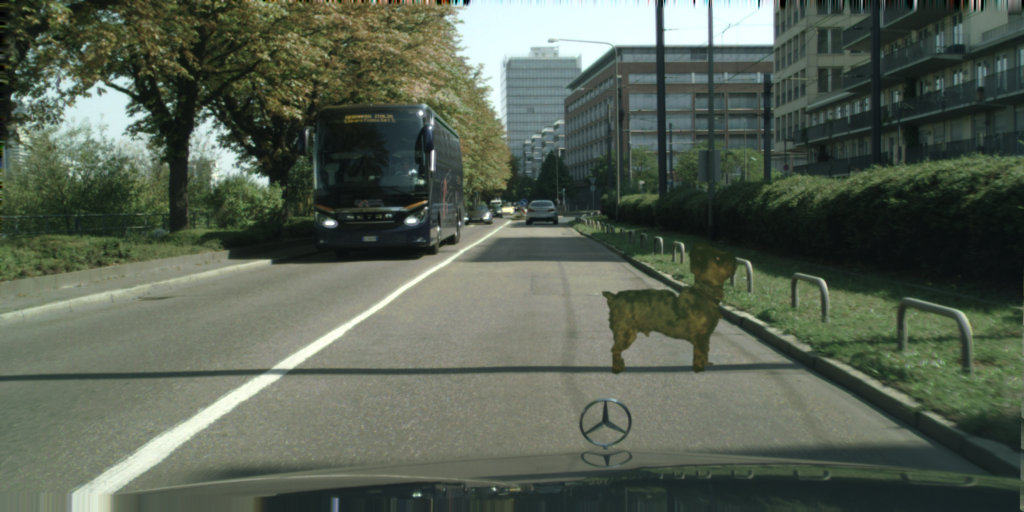}
            \caption[]%
            {{Image input}}    
            \label{}
        \end{subfigure}
        \begin{subfigure}[b]{0.22\textwidth}  
            \centering 
            \includegraphics[width=\textwidth]{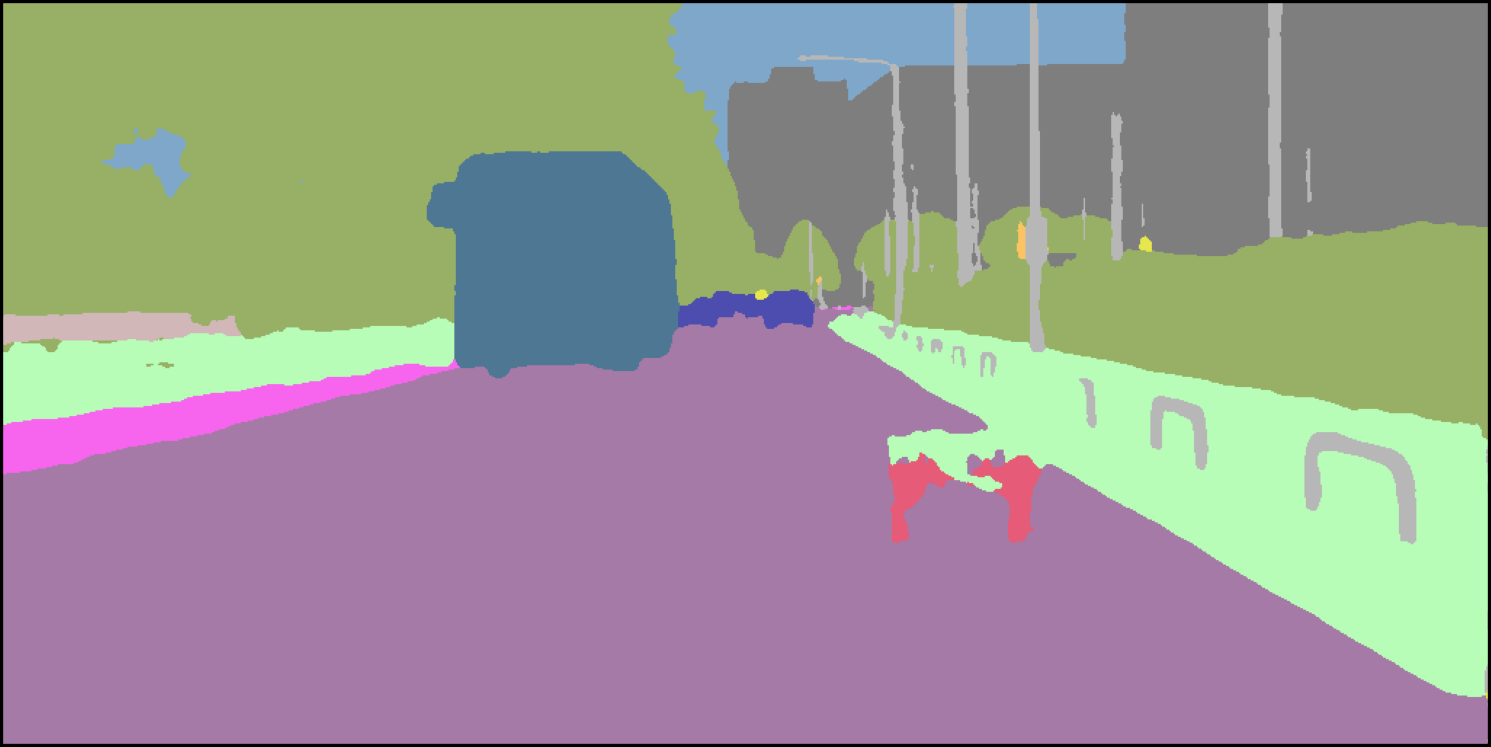}
            \caption[]%
            {{Segmented image}}    
            \label{}
        \end{subfigure}
        \begin{subfigure}[b]{0.22\textwidth}   
            \centering 
            \includegraphics[width=\textwidth]{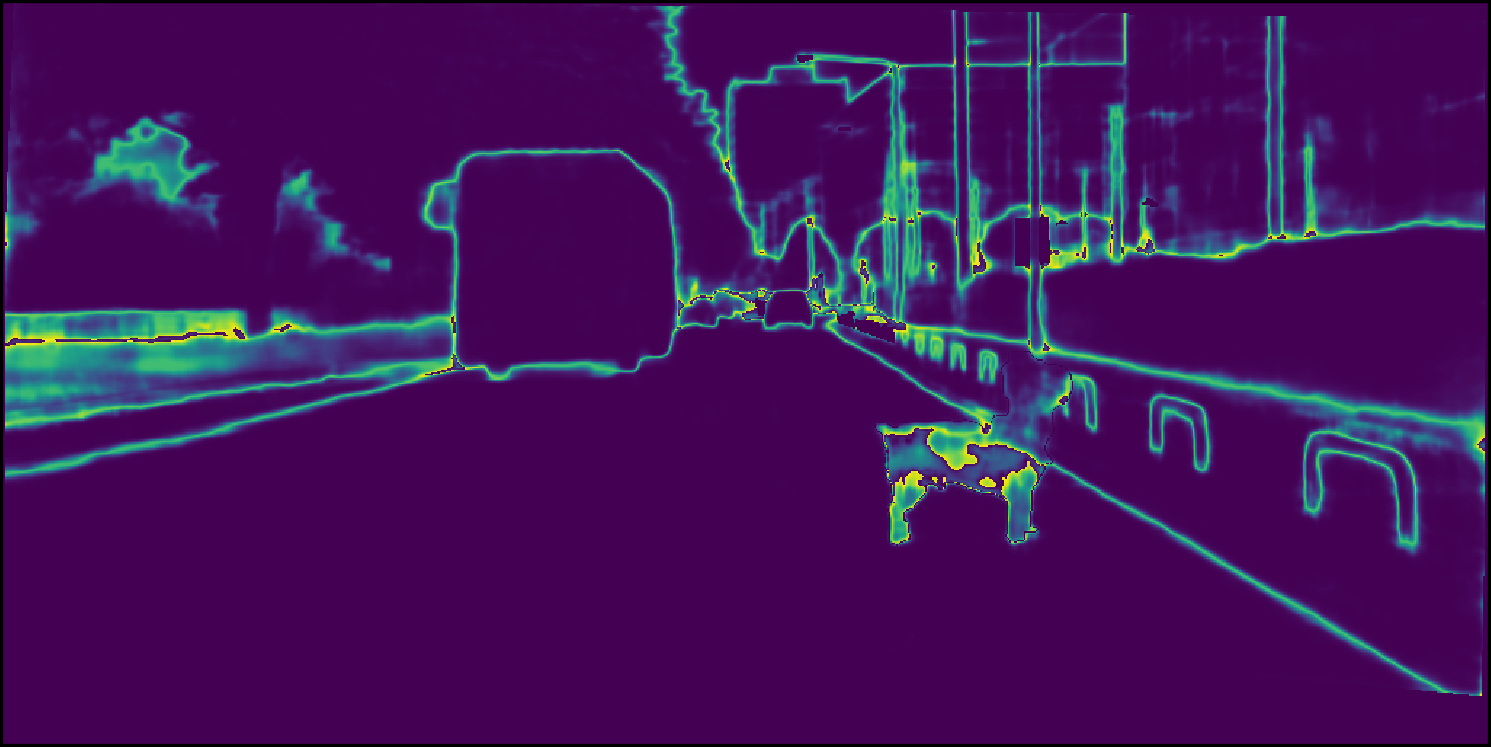}
            \caption[]%
            {{Entropy map}}    
            \label{}
        \end{subfigure}
        \begin{subfigure}[b]{0.22\textwidth}   
            \centering 
            \includegraphics[width=\textwidth]{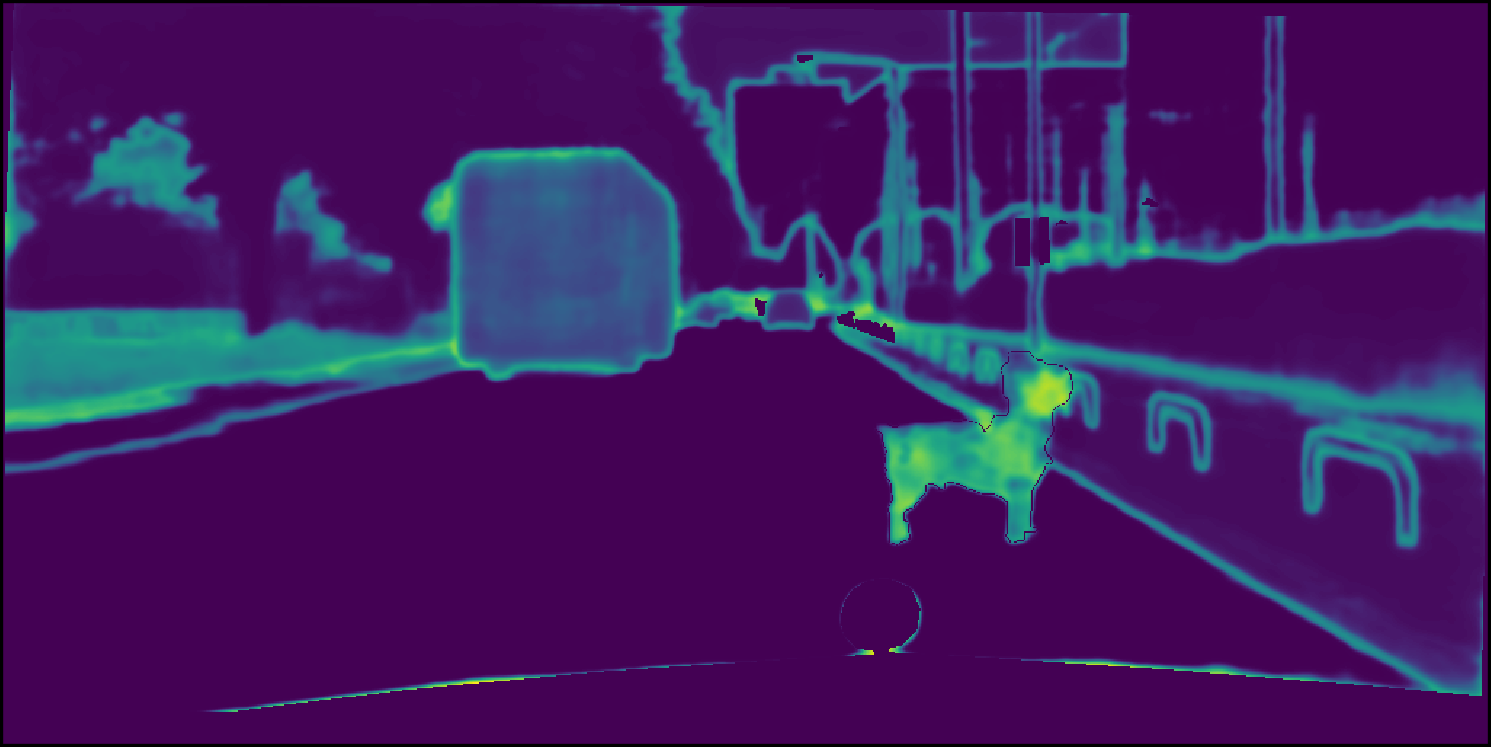}
            \caption[]%
            {{OOD estimate}}    
            \label{}
        \end{subfigure}
        \begin{subfigure}{0.022\textwidth} 
           \includegraphics[width=\linewidth, clip,trim=0in 0in 0.03in 0.0in]{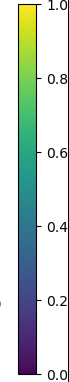}
           \label{fig:subim2}
       \end{subfigure}
        \caption[]
        {\small Example input-output for MEMOS from Fishyscapes dataset. Input image (a) passes through a segmentation net, yielding segmentation output (b) and prediction entropy (c). Images (b) and (c) are combined channel-wise as input to the metacognitive net, which outputs (d): per-pixel predictions of OOD vs ID. Low entropy is blue, high entropy is yellow.}
        \label{fig:metacog_imaged}
        \vspace{-0.7cm}
    \end{figure}

We present a framework that detects OOD instances without relying on any OOD data during training, by leveraging measures of dispersion of well-calibrated segmentation networks in a metacognitive network module. 
We make further use of the available in-distribution (ID) data by transforming portions of ID images into unrecognizable classes, creating synthetic OOD data situated in an ID context. 
The combined ID and synthetic OOD data is used to improve the entropy calibration of segmentation models through maximum entropy training \cite{pereyra2017regularizing} by 1) tightening the boundary around low-entropy predictions, and 2) reducing the network's confidence on identifying synthetic OOD data, which primes the network to make high-entropy predictions more frequently when encountering actual OOD data. 
Our metacognitive network then uses these predictions to make final context-supported decisions on which pixels belong to OOD class instances 
(Figure~\ref{fig:metacog_imaged}).  
We refer to our framework as Maximum-Entropy Metacognitive OOD Segmentation (MEMOS). 


Our key contributions include:
\begin{itemize}[noitemsep,topsep=0pt,parsep=0pt,partopsep=0pt,leftmargin=1em]
    \item We develop a metacognitive network module that leverages the predicted class, entropy, and spatial context to generate pixel-wise OOD detection for segmentation. 
    \item We propose a method for generating synthetic OOD data from ID data that improves the entropy calibration of segmentation models through maximum entropy training.
    \item 
    The full MEMOS framework reliably identifies novel instances in real-time (30-40 Hz), and achieves up to a 75\% increase in performance  over comparable methods.  
\end{itemize}

\section{BACKGROUND AND RELATED WORK}
\label{sec:related work}


\textbf{OOD Detection for Segmentation}~~
The most common approach to enable segmentation models to detect OOD inputs is to provide sample OOD data to the training process and optimize explicitly to detect these OOD samples~\cite{bevandic2019simultaneous,chan2021entropy,bevandic2021dense,du2022vos}---an approach that can bias the model, as discussed previously.
Other methods reason about the uncertainty of the prediction by using handcrafted metrics derived from the segmentation model's output confidences~\cite{jung2021standardized,chan2021entropy,hendrycks2022scaling}, by combining these confidences with other approaches~\cite{di2021pixel}, by using Bayesian Neural Networks~\cite{kendall2015bayesian}, or by using ensembles~\cite{lakshminarayanan2017simple}. 
%
Reconstruction-based approaches attempt to reconstruct the input image from intermediate representations or the final semantic segmentation, intuiting that regions that are difficult to reconstruct are most likely OOD~\cite{creusot2015real,lis2019detecting,lis2020detecting,ohgushi2020road,vojir2021road,di2021pixel}. Our approach is most similar to Di Biase, et al.~\cite{di2021pixel}, in that we use a network to reason about the segmentation network's confidence, but we do not require an expensive secondary reconstruction-based pipeline. 
%
%
%
Other approaches reason about temporal information~\cite{huang2018efficient}, in contrast to our method that only requires a single frame.
A survey of different anomaly detection approaches is available from Bogdoll et al.~\cite{bogdoll2022anomaly}.





\textbf{Maximum Entropy}~~
Many methods attempt to exploit the entropy of neural network outputs for OOD detection.
Prior work has regularized the network by penalizing low entropy scores~\cite{pereyra2017regularizing} or training to maximize the entropy on known OOD images~\cite{chan2021entropy,pinto2021mix}. We take a similar maximum entropy approach to increase the entropy of the base segmentation net on OOD data, and thus refine the inputs to our metacognitive network, although we leverage synthetic generation to remove the dependency on 
known OOD training samples. 



\textbf{Synthetic OOD Data Generation}~~
Throughout this paper, we distinguish between three types of data:  ID, OOD, and synthetic OOD. We define {\em ID  data} as all data available to an agent for its segmentation task at training, {\em OOD data} as any data the agent will encounter after deployment beyond the ID classes, and {\em synthetic OOD data} as generated data whose use in training improves OOD detection. 
We make these distinctions to motivate our means of refining models over a limited ID training set, which we discuss below and detail in Section~\ref{sec-synth-OOD}.

Generating synthetic OOD data is a common approach for training OOD detection when OOD data is inaccessible. It is typically used for image classification, but is underexplored for segmentation. Several approaches generate synthetic OOD data for image classification by interpolating between examples in the ID dataset~\cite{zhang2017mixup,pinto2021mix,pinto2022regmixup}, or performing image transformations to corrupt ID data~\cite{hebbalaguppe2023novel}. These techniques allow for easily generating large quantities of data likely to be outside of the training data distribution, but the generated images may not be visually realistic. Further, it is unclear how to apply the interpolation techniques to segmentation, where instances of different classes are completely different shapes. Such an approach would have to address how to select instance pairs for interpolation, and how to realistically compose full images from corrupted class instances together with ID data. 
Generative models have also been used to produce synthetic OOD data for image classification~\cite{lee2017training} 
by training a generator model capable of producing fully-artificial OOD images, which can be time- and resource-intensive. Further, adapting generative approaches for classification to semantic segmentation problems is non-trivial. 

In contrast to the above methods, our approach generates synthetic OOD data by heavily blurring a random subset of ID class instances into something unrecognizable as the ID data (similar to Hebbalaguppe et al.~\cite{hebbalaguppe2023novel}), creating mixed synthetic OOD and ID training images. This does not require training a resource-intensive generator, and yields synthetic OOD data that fit directly among ID data as context.



\section{Approach}
\label{sec:approach}

\begin{figure*}[t]
\centering
\includegraphics[width=\textwidth]{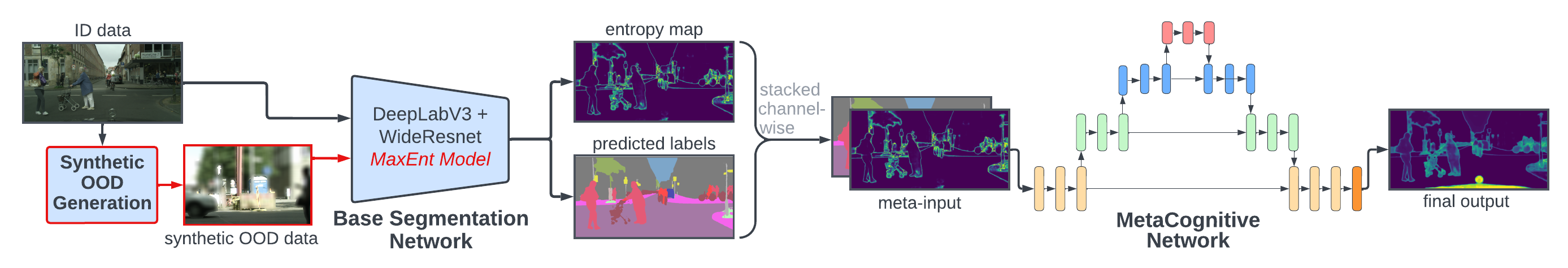}
\caption{An overview of our MEMOS framework. Synthetic OOD data is generated and used to fine-tune the base segmentation network via maximum entropy training. Predicted labels and the entropy map are stacked as channels for input into the metacognitive network, which generates a final uncertainty mask for OOD detection.
}
\vspace{-0.3cm}
\label{fig:framework_arch}
\end{figure*}


Our MEMOS framework is comprised of two components: 1)~a standard base segmentation network fine-tuned with maximum entropy training (detailed in Section \ref{sec:me-seg}) over ID and generated synthetic OOD data (see Section \ref{sec-synth-OOD}), and 2)~a metacognitive network module (Section~\ref{sec:mc-net}) that reasons about the base model's predictions in order to identify OOD instances. The full framework is shown in Figure~\ref{fig:framework_arch}.
The core of the approach is the novel metacognitive network, which we design as a modular component that can re-use training data from the base model. This module sits on top of the fine-tuned base model, using its class predictions, their dispersion, and their spatial relationships to each other to generate a binary mask that grades the quality and correctness of the base segmentation net's predictions.

Our key insight is that semantic segmentation networks are originally trained only for segmentation, not OOD detection. By adding a lightweight metacognitive network that makes improved quality judgements, we can use these quality judgements directly for OOD detection. Prior work has established that the entropy of network predictions is highly indicative of uncertainty and can be relied upon for OOD detection~\cite{chan2021entropy,pinto2021mix, pinto2022regmixup}, and our metacognitive network offers improved uncertainty judgements. Additionally, the metacognitive network relies on entropy as an input, and thus improves further as the entropy of the segmentation network's outputs are better calibrated on both ID and OOD data. To achieve this improvement in entropy estimates, we fine-tune the base segmentation network using maximum entropy training over a dataset composed of ID data and our generated synthetic OOD data. We detail the fine-tuning process and the resulting  MaxEnt segmentation network in Section~\ref{sec:me-seg}, and the supporting synthetic OOD data generation in Section~\ref{sec-synth-OOD}. 

\subsection{The Metacognitive Network}
\label{sec:mc-net}


Most approaches make OOD detections by simply thresholding uncertainty measures \cite{hendrycks2016baseline, liang2017enhancing} output either by a single network or an ensemble. However, such simple thresholding fails to take contextual cues into account. While an end-to-end segmentation model trained for OOD detection has the necessary context to produce accurate uncertainty measures, this does not always happen in practice. For example, the OOD dog in Figure~\ref{fig:metacog_imaged}(c) is difficult to discern from simple pixel-wise thresholding alone. However, all information needed to make this decision is present in the segmentation network's output   (Figures~\ref{fig:metacog_imaged}(b) and (c)), suggesting that this is a failure of it considering its own segmentation confidence, and suggesting a metacognitive approach would enable more effective OOD detection (Figure~\ref{fig:metacog_imaged}(d)). 



The metacognitive network module, which appends to any base segmentation network, computes a quantitative uncertainty estimate for each prediction, which can then be used for OOD detection. We train the metacognitive network 
using the binary cross entropy loss function $\mathcal{L}_{\mathrm{bce}}$ as follows:
\begin{equation}
     \label{equ:MetaLoss}
    \argmin_\phi \sum_{x \in D_{\mathrm{id}}} \mathcal{L}_{\mathrm{bce}}(\phi, g(x)) \enspace .
\end{equation}
We generate the metacognitive input $g(x)$ by stacking the pixel-wise entropy and predicted class channel-wise, keeping the original image structure to maintain spatial context:
\begin{equation}
    \label{equ:Metainput}
    g(x) = \argmax_{y\in Classes}f_{seg}(\theta, x) ^\frown \Omega_{ent}(\theta, x) \enspace ,
\end{equation} where $f_{seg}$ is a base segmentation network with  parameters $\theta$, $\phi$ are the parameters for the metacognitive network, $D_{id}$ is the ID training data, and  $\Omega_{\mathrm{ent}}(\theta, x)$ is the entropy of the conditional distribution $p_{\theta}(\mathbf{y}|\mathbf{x})$ produced by a network with parameters $\theta$ over classes $\mathbf{y}$ for the input $\mathbf{x}$, defined as
\begin{equation}
    \Omega_{\mathrm{ent}}(\theta, x) = - \!\!\!\!\!\sum_{i \in \mathrm{Classes}}\!\!\! p_{\theta}(\mathbf{y}_{i}| \mathbf{x}) \log p_{\theta}(\mathbf{y}_{i}| \mathbf{x}) \enspace .
    \label{eq:entropy}
\end{equation}

The intuition behind this input structure is that reasoning about the distribution of certainty in the model's predictions is a strong indicator of novelty~\cite{chan2021entropy}, and including the predicted classes allows the metacognitive network to condition its reasoning based on different commonly observed arrangements of classes. For example, we would expect higher entropy on the border of similar correctly classified objects, such as grass bordering a bush, but would expect very low entropy on the borders of easily distinguished classes, such as grass bordering a building. Additionally, we do not include any of the original image data to prevent overfitting while also keeping the metacognitive network space and time efficient. As it computes its input from class predictions that essentially come from a black box, the metacognitive network does not depend on the architecture of the segmentation model and can be used with any compatible segmentation pipeline.

We use a U-Net architecture~\cite{RFB15a} for our metacognitive network (Figure~\ref{fig:framework_arch}) to enable the network to reason about both local features and the larger scale image context.
We use a subset of the training data and its corresponding predictions from the segmentation network to train the metacognitive network. We compute a target label as a binary mask indicating correct vs. incorrect segmentation network predictions, where 0 indicates a correct prediction and 1 indicates an incorrect prediction. 
The metacognitive network predicts a soft mask (Figure \ref{fig:metacog_imaged}(d)), consisting of values in $[0,1]$ for each pixel in the corresponding input image. An estimate closer to $1$ indicates a poor, highly uncertain prediction; as the value moves closer to $0$, the metacognitive network is more certain about the segmentation prediction. The network is trained using a binary cross-entropy loss to learn a function of entropy, correlated with each predicted class, aided by the context of neighbouring pixels, to determine the quality of the segmentation prediction. Pixels with high uncertainty can then be identified as OOD data.

OOD detection could likely be improved further by filtering out contiguous detected regions under a size threshold---a post-processing step. We do not evaluate post-processing approaches as shown in other work~\cite{chan2021entropy} to reduce the dependent variables evaluated in the scope of this paper, and leave it to future work. However, we do expect many post-processing approaches could benefit this framework, as the metacognitive network's inclusion of context from neighboring pixel entropy and class labels enables it to detect more contiguous regions, as seen in 
Figures~\ref{fig:metacog_imaged}(c) and (d). This can be both a benefit and a hindrance: the now-contiguous dog is correctly labeled OOD and the contiguous bus mirror is incorrectly labeled OOD. However, our ablation studies in Section \ref{sec:result} show this is more beneficial overall.

\subsection{Maximum Entropy Segmentation Model}
\label{sec:me-seg}


\begin{figure}[t]
 \centering
    \begin{subfigure}[b]{0.23\textwidth}
        \centering
        \includegraphics[width=\textwidth]{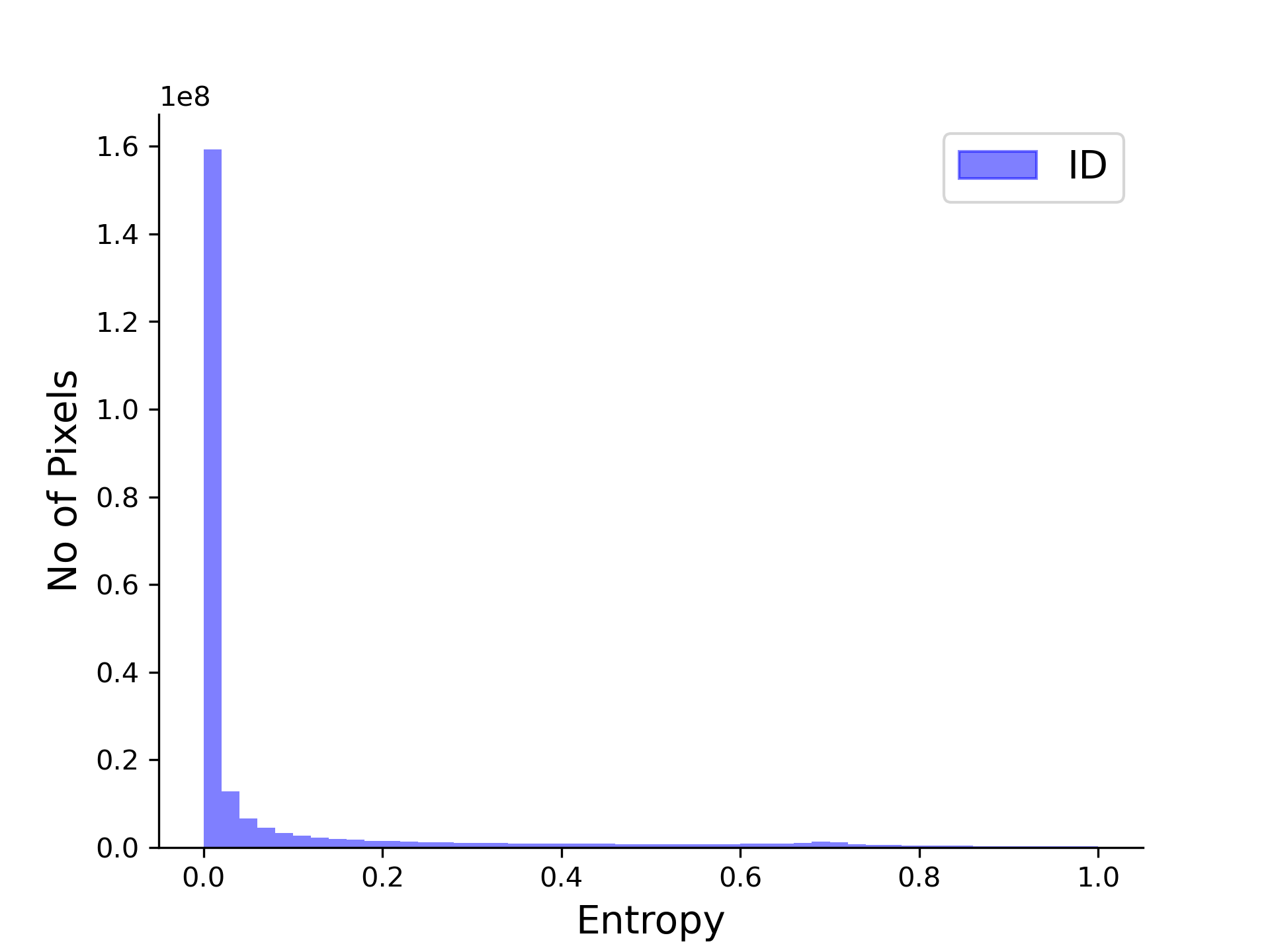}
        \caption[]%
        {{\tiny base model, ID data}}    
        \label{}
    \end{subfigure}
    \begin{subfigure}[b]{0.23\textwidth}   
        \centering 
        \includegraphics[width=\textwidth]{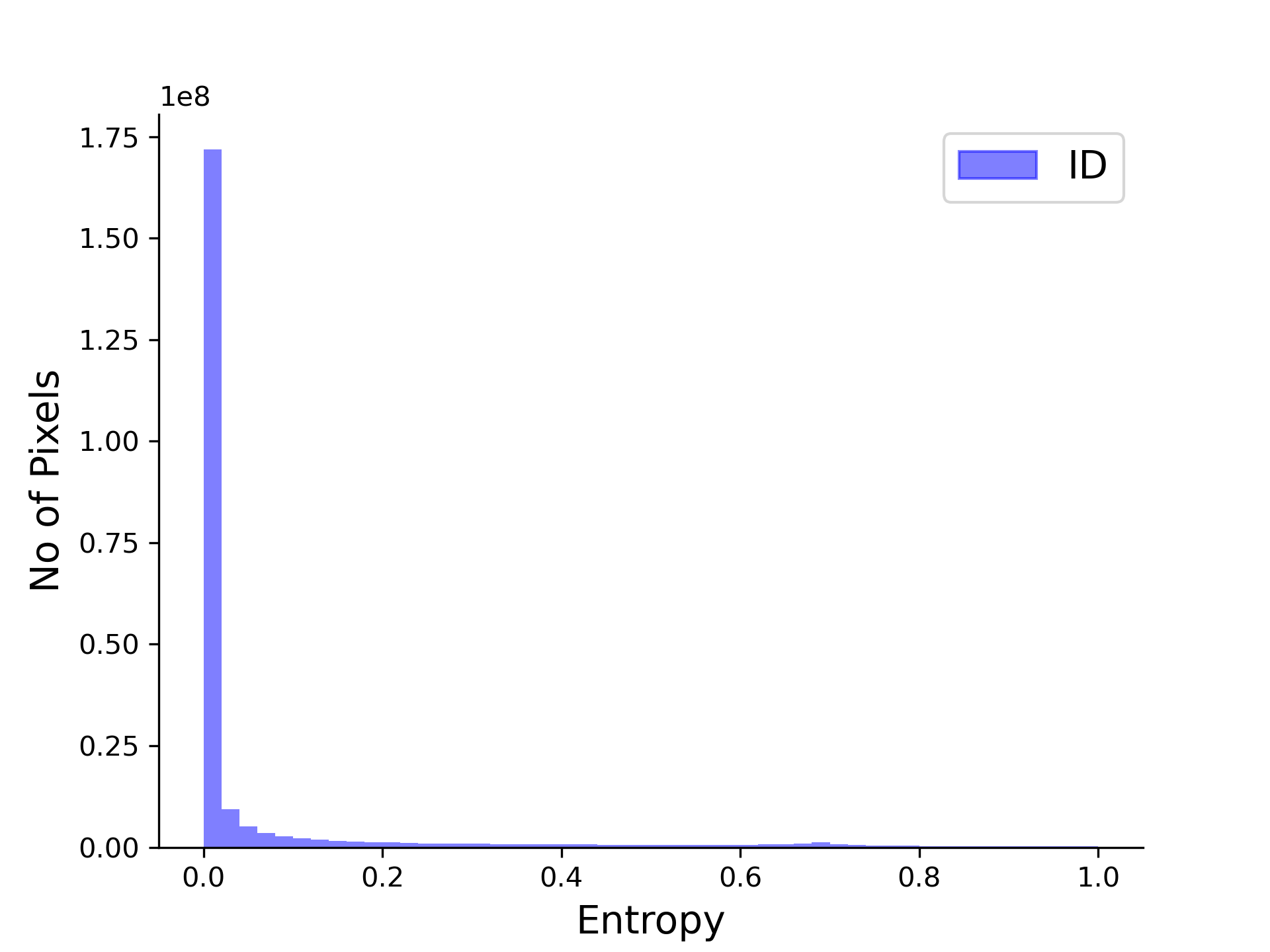}
        \caption[]%
        {{\tiny  MaxEnt model, ID data}}    
        \label{}
    \end{subfigure}
    
    \begin{subfigure}[b]{0.23\textwidth}  
        \centering 
        \includegraphics[width=\textwidth]{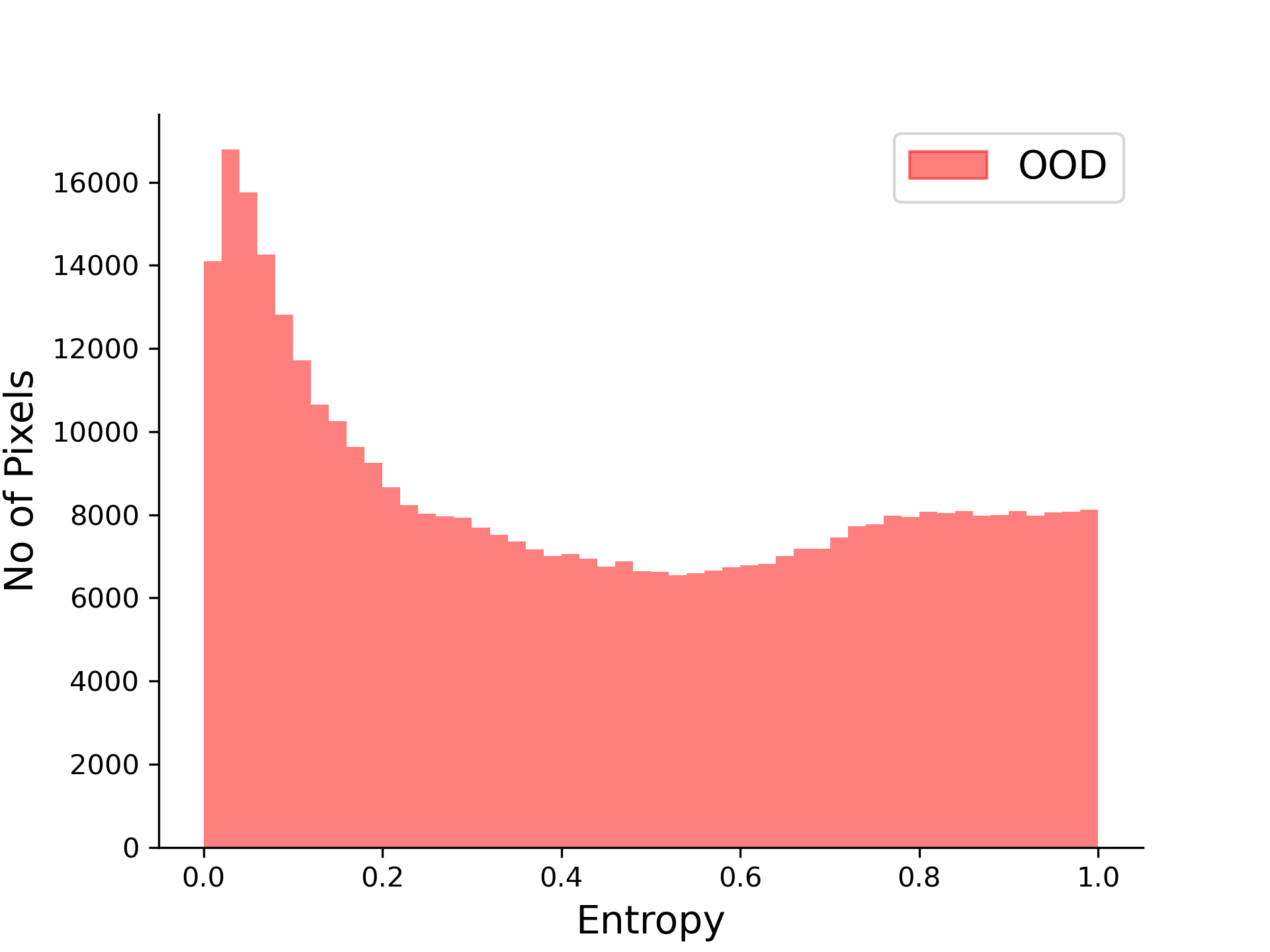}
        \caption[]%
        {{\tiny base model, OOD data}}    
        \label{}
    \end{subfigure}
    \begin{subfigure}[b]{0.23\textwidth}   
        \centering 
        \includegraphics[width=\textwidth]{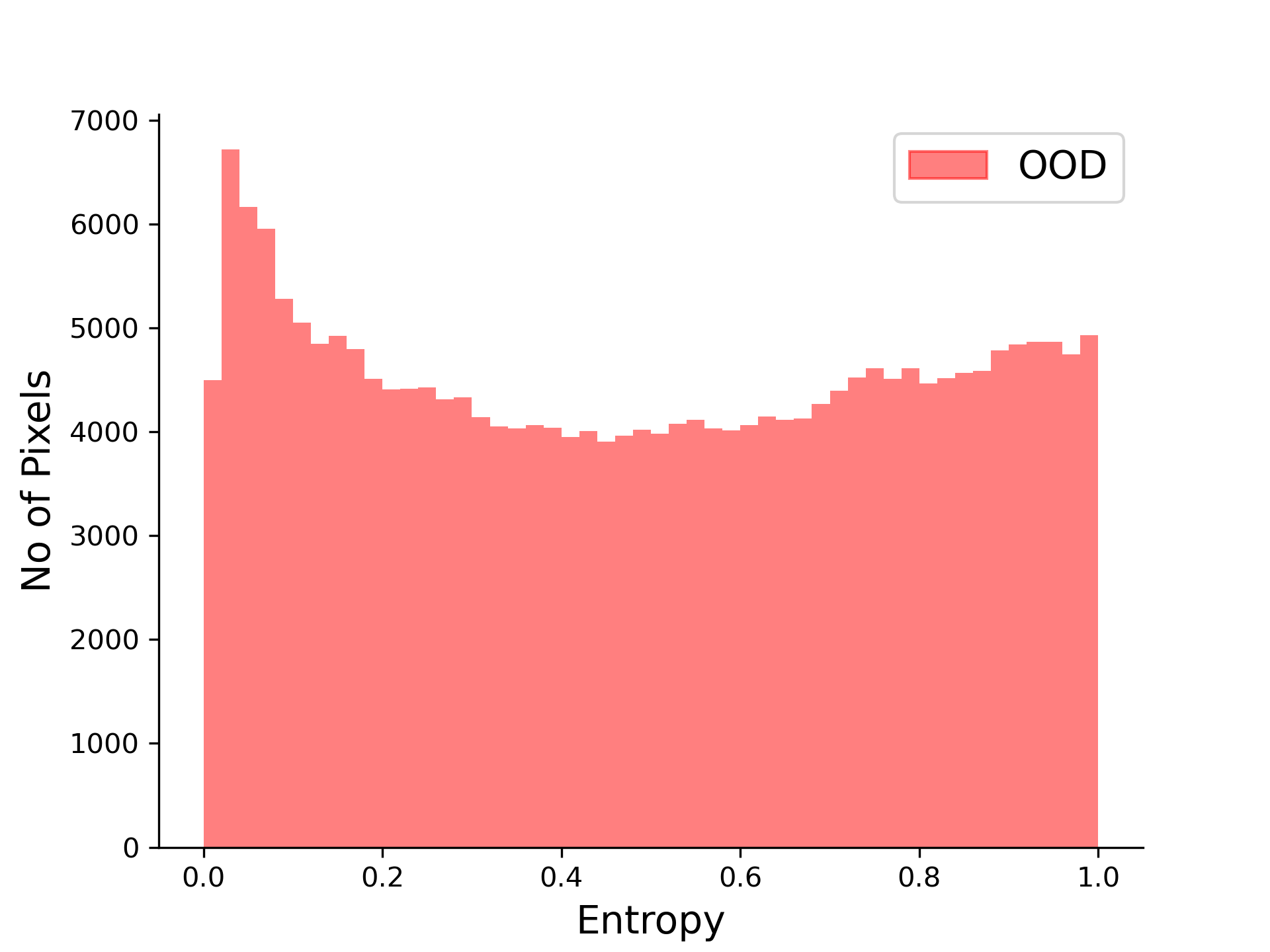}
        \caption[]%
        {{\tiny MaxEnt model, OOD data}}    
        \label{}
    \end{subfigure}
    \caption[]
    {Histograms of entropy distribution across ID and OOD data for the base segmentation model and our MaxEnt segmentation model.}
    \label{fig:ent_hist}
\end{figure}

\begin{figure}[t]
\centering
   \begin{subfigure}{0.15\textwidth}
    \includegraphics[width=\linewidth,clip,trim=30px 80px 30px 28px]{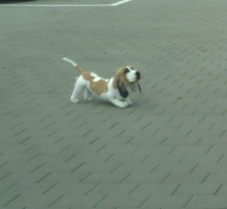}
       \caption{OOD Instance}
       \label{fig:subim1}
   \end{subfigure}
   \begin{subfigure}{0.15\textwidth}
       \includegraphics[width=\linewidth,clip,trim=0.65in .6in .98in 1.5in]{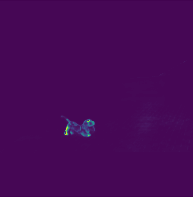}
       \caption{Base model}
       \label{fig:subim2}
   \end{subfigure}
   \begin{subfigure}{0.15\textwidth}
       \includegraphics[width=\linewidth,clip,trim=0.65in .6in .98in 1.5in]{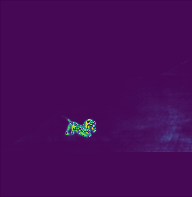}
       \caption{MaxEnt model}
       \label{fig:subim3}
   \end{subfigure}
    \begin{subfigure}{0.25\textwidth}
       \includegraphics[width=\linewidth]{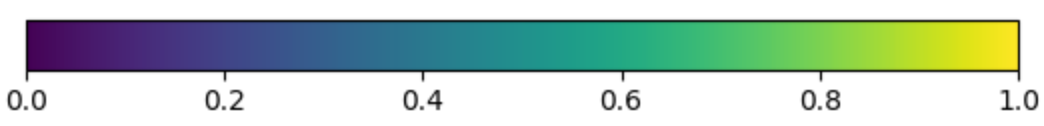}
   \end{subfigure}
   \caption{Entropy heatmap of (a) an OOD instance for (b) the standard segmentation model and (c) the segmentation model with maximum entropy fine-tuning. Yellow pixels indicate high entropy and uncertainty.}
   \label{fig:image2}
\end{figure}

Large DNNs are known to be overconfident about their predictions \cite{guo2017calibration, mukhoti2020calibrating}, resulting in unreliable uncertainty estimates for downstream components. This is a concern for OOD detection through our metacognitive network, since the base segmentation network typically underestimates the entropy for OOD data due to overconfidence. The result is that the base network is skewed towards low-entropy predictions, even on OOD data, as shown in Figure \ref{fig:ent_hist}(c). To alleviate this issue, we calibrate the base segmentation model to generate high-entropy predictions for OOD instances and low-entropy predictions for ID instances by fine tuning it using synthetic OOD data; we refer to this as the MaxEnt segmentation model.
Figure \ref{fig:ent_hist}(c) and (d) show that our MaxEnt model does produce a broader entropy distribution over OOD data from the same samples, including more high-entropy predictions, when compared to the original segmentation network. Additionally, we see improvement in the frequency of low-entropy predictions for ID samples in comparison to the base segmentation model (Figures \ref{fig:ent_hist}(a) and (b)). Figure \ref{fig:image2} shows a qualitative example of the difference between the base and  calibrated MaxEnt models, the latter of which has higher entropy predictions for the OOD puppy.

 To learn the MaxEnt segmentation model, we add a maximum entropy regularizer \cite{pereyra2017regularizing} to the standard cross entropy loss $\mathcal{L}_{\mathrm{ce}}$. The resulting optimization is given by
\begin{equation}
     \label{equ:MaxEntLoss}
    \argmin_\theta \sum_{x \in D_{\mathrm{id}}} \mathcal{L}_{\mathrm{ce}}(\theta, x) -  {\color{gray}\underbrace{ \color{black} \lambda\!\!\!\sum_{x \in D_{\mathrm{ood}}}\!\!\!\Omega_{\mathrm{ent}}(\theta, x)}_{\mbox{\color{gray} \hspace{-2em}\small MaxEnt regularizer \hspace{-2em}}}} \enspace ,
\end{equation}
where  
 $D_{\mathrm{id}}$ represents the ID training data, $D_{\mathrm{ood}}$ represents the synthetic OOD training data described in Section \ref{sec-synth-OOD}, $\lambda$ is the regularization parameter, and
$\Omega_{\mathrm{ent}}$ is the entropy as defined in Equation~\ref{eq:entropy}. The regularization term encourages the segmentation network to maximise entropy for the predictions when input $\mathbf{x}$ is synthetic OOD data. 
Training the segmentation by optimizing  Equation~\ref{equ:MaxEntLoss} calibrates the network to make low-entropy predictions for ID samples and high-entropy predictions for synthetic OOD samples, which should translate to actual OOD samples at deployment.  


\subsection{Synthetic OOD Data Generation}

\label{sec-synth-OOD}
The MaxEnt approach 
raises an important question, namely how to access OOD training data for the \textit{MaxEnt regularizer} to significantly effect the model's performance? Some approaches train using a predetermined set of known OOD data~\cite{chan2021entropy}, but we treat the separation of ID and OOD data as inviolable, as we define OOD data as unknown pre-deployment. One could divide the training data to create a subset of OOD classes~\cite{vyas2018out}, but this would reduce the amount of ID training data and impact ID performance. We circumvent this issue by using a data augmentation approach based on transforming ID images for classification~\cite{zhang2017mixup,pinto2021mix,pinto2022regmixup,hebbalaguppe2023novel}. To our knowledge, this is the first approach that generates synthetic OOD data from ID data for segmentation.
Specifically, we create a randomly sampled subset of the training data $D_\mathrm{sub} \subset D_{\mathrm{id}}$, and a subset of the ID classes, $C_\mathrm{sub}$. For every image in $D_\mathrm{sub}$, we apply a Gaussian blur on all pixels belonging to classes not in $C_\mathrm{sub}$. $D_\mathrm{sub}$ is added to the train set as additional data and the Gaussian blurred pixels in every image serve as synthetically generated OOD data, $D_\mathrm{ood}$. 

This data augmentation approach has a few advantages. First, it ensures our synthetic \OOD data is directly based on natural images, and produces objects of a similar size, resolution, and intensity. Second, it keeps some ID data belonging to $C_\mathrm{sub}$ in each training image to provide context for the synthetic \OOD instances. Last, it also provides training data for ID classes that are in the presence of synthetic \OOD instances, simulating a condition we expect to occur after the model is trained and deployed. We show that this data augmentation approach successfully increases entropy for \OOD samples in Figure \ref{fig:ent_hist}, an example of which is shown in Figure \ref{fig:image2}, and we evaluate its effect on OOD performance via ablation studies in Section \ref{sec:result}.
	

\section{EXPERIMENTS AND RESULTS}
\label{sec:result}

Evaluating the performance of various OOD detection methods requires both ID and OOD datasets. 
We use Cityscapes~\cite{cordts2016cityscapes} as our ID data, which consists of street scenes from 50 different cities; we train both our MaxEnt and metacognitive networks using its 2,975 image trainset. We evaluate OOD detection on the following established OOD detection for semantic segmentation benchmarks from the Lost and Found~\cite{pinggera2016lost} and Fishyscapes~\cite{blum2021fishyscapes} benchmarks, which comprise of road and street scenes with anomalies not present in Cityscapes: the test split of Lost and Found dataset, 
the Fishyscapes Static dataset, and the Fishyscapes (FS) Lost and Found dataset. Wherever possible, we use the same base segmentation model for comparison across experiment conditions. Specifically, we use the state-of-the-art DeepLabV3+ with a WideResnet38 backbone using pretrained weights~\cite{zhu2019improving}. For further details, experiment code, models, and hyperparameters, see our repository\footnote{\url{https://github.com/meghna30/metacognitive_segmentation/tree/main}} .

\subsection{MaxEnt Model and Metacognitive Training Details} 
We begin by training the MaxEnt model (Section~\ref{sec:me-seg}) using the Cityscapes trainset as $D_\mathrm{id}$ along with a synthetic OOD dataset $D_\mathrm{ood}$. To generate $D_\mathrm{ood}$, we randomly sample $500$ images from the Cityscapes trainset, and select 12 classes ($|C_\mathrm{sub}| = 12$) to remain in-distribution. The classes were selected randomly to avoid bias. We found that selecting approximately half of the classes to generate synthetic OOD data gave the best performance, and hypothesize this is effective as it provides an even balance of contextualized synthetic OOD data in the individual images.

Next, we train the metacognitive network, as described in Section~\ref{sec:mc-net}, constructing a trainset by randomly sampling 500 images and their corresponding ground truth labels from $D_\mathrm{id}$. We compute the final training labels from their corresponding ground truth labels and the output of the previously trained MaxEnt network. 

During evaluation, we pass each test sample from the OOD datasets through the MaxEnt segmentation model, construct the input $g(x)$, and pass $g(x)$ through the metacognitive network to predict the final detection mask, as shown in Figure~\ref{fig:framework_arch}.

\subsection{Evaluation Procedure}

\begin{table*}[t]
    \begin{minipage}[b]{0.75\linewidth}
    \setlength{\tabcolsep}{3.5pt}
    \centering
    \captionsetup{width=\linewidth}
    \caption{Performance of OOD detection methods across Lost and Found and Fishyscapes benchmarks. Methods below the horizontal line require access to OOD data during training.}
    \label{tab:results1}
     \begin{tabular}{r|c|c|ll|ll|ll}
        \hline
            OOD Detection & OOD & \multicolumn{1}{c|}{Val} & \multicolumn{2}{c|}{Lost and Found} & \multicolumn{2}{c|}{FS Lost and Found} &  \multicolumn{2}{c}{Fishyscapes - Static} \\
        Method  & Data? &   mIoU & AUPRC & FPR-95 & AUPRC & FPR-95 & AUPRC & FPR-95\\
        \hline\hline
        Softmax     && 0.89\scriptsize$\pm$.008 & 0.26\scriptsize$\pm$.002 & 0.17\scriptsize$\pm$.023 & 0.05\scriptsize$\pm$.012 & 0.36\scriptsize$\pm$.055 & 0.18\scriptsize$\pm$.045 & 0.19\scriptsize$\pm$.016\\ 
        Entropy      &    & 0.89\scriptsize$\pm$.008   & 0.44\scriptsize$\pm$.001 & 0.22\scriptsize$\pm$.102 & 0.13\scriptsize$\pm$.031 & 0.33\scriptsize$\pm$.061 & 0.35\scriptsize$\pm$.034 & 0.18\scriptsize$\pm$.018\\        
        Ensemble     &      & 0.88\scriptsize$\pm$.000  & 0.07\scriptsize$\pm$.005 & 0.26\scriptsize$\pm$.087 & 0.02\scriptsize$\pm$.302 & 0.29\scriptsize$\pm$.017  & 0.32\scriptsize$\pm$.021 & 0.16\scriptsize$\pm$.006 \\ 
        Learned Density &   & 0.80  & --- & --- & 0.04 & 0.47  & 0.62 & 0.17 \\
        MEMOS (Ours)    &    & 0.87\scriptsize$\pm$.000 & \textbf{0.70\scriptsize$\mathbf{\pm}$.012} & 0.12\scriptsize$\pm$.037& \textbf{0.23\scriptsize$\mathbf{\pm}$.005} & 0.46\scriptsize$\pm$.172 & \textbf{0.65\scriptsize$\mathbf{\pm}$.045} & 0.35\scriptsize$\pm$.126\\ 
        \hline
        ODIN & \checkmark & 0.89\scriptsize$\pm$.008 &  0.56\scriptsize$\pm$.008 & 0.12\scriptsize$\pm$.009& 0.15\scriptsize$\pm$.014 & 0.27\scriptsize$\pm$.11 & 0.13\scriptsize$\pm$.03 & 0.49\scriptsize$\pm$.016\\ 
        EM-Coco  & \checkmark & 0.89 & 0.76 & 0.095 & 0.41 & 0.37 & 0.81 & 0.094\\
        EM-Coco\&Meta  & \checkmark & 0.89 & \textbf{0.79} & 0.009 & \textbf{0.43} & 0.43 & \textbf{0.84} & 0.11\\
        \hline
        \end{tabular}
    \end{minipage}%
    \begin{minipage}[b]{0.23\linewidth}
    \centering
    \captionsetup{width=\linewidth}
    \caption{Inference time}
    \label{tab:infer_time}
    \begin{tabular}{r|l}
        \hline
             Method & Inf. Time (ms) \\
        \hline \hline
        Synboost     & \cellcolor{red!50} $1,055.5$ \\ 
        Add'l Conv. & \cellcolor{red!38.7} $816.9$\\
        DeepLabV3      & \cellcolor{red!1.2} $24.5$ \\
        Softmax     & \cellcolor{red!1.2} $24.5$ \\
        Entropy      & \cellcolor{red!1.2} $24.5$ \\
        Ensemble              & \cellcolor{red!3.5} $24.5\times3$ \\
        ODIN           & \cellcolor{red!29.4} $24.5+595.7$ \\
        EM-COCO     & \cellcolor{red!1.2} $24.5$  \\
        MEMOS (Ours)       & \cellcolor{red!1.5} $24.5+6.4$ \\
      \hline
        \end{tabular}
    \end{minipage}
\end{table*}

\begin{table*}[t]
    \begin{minipage}[h]{\linewidth}
    \centering
    \captionsetup{width=\linewidth}
    \caption{Ablation studies across Lost and Found and Fishyscapes benchmarks }
    \label{tab:ablation1}
    \begin{tabular}{r|c|ll|ll|ll}
        \hline
            & \multicolumn{1}{c|}{Val} & \multicolumn{2}{c|}{Lost and Found} &  \multicolumn{2}{c|}{FS Lost and Found} & 
            \multicolumn{2}{c}{Fishyscapes - Static}\\
         OOD Detection &  mIoU & AUPRC & FPR-95 & AUPRC & FPR-95 & AUPRC & FPR-95\\
        \hline \hline
        Entropy           & 0.89\scriptsize$\pm$.008  & 0.44\scriptsize$\pm$.001 & 0.22\scriptsize$\pm$.102 & 0.13\scriptsize$\pm$.031 & 0.33\scriptsize$\pm$.061 & 0.35\scriptsize$\pm$.034 & 0.18\scriptsize$\pm$.018\\ 
        Add'l Conv.    & 0.80\scriptsize$\pm$.008  & 0.45\scriptsize$\pm$0.021 & 0.30\scriptsize$\pm$0.016 & 0.12\scriptsize$\pm$0.063 & 0.41\scriptsize$\pm$0.057 & 0.22\scriptsize$\pm$0.034 & 0.25\scriptsize$\pm$0.092\\ 
        Metacognitive-Only          & 0.85\scriptsize$\pm$.009  & 0.48\scriptsize$\pm$.014 & 0.15\scriptsize$\pm$.031 & 0.13\scriptsize$\pm$.033 & 0.55\scriptsize$\pm$.1 & 0.39\scriptsize$\pm$.043 & 0.26\scriptsize$\pm$.062\\ 
        MaxEnt          & 0.90\scriptsize$\pm$.000 & 0.64\scriptsize$\pm$.0.005 & 0.28\scriptsize$\pm$.066  & 0.22\scriptsize$\pm$.045 & 0.24\scriptsize$\pm$.013 & 0.61\scriptsize$\pm$.060 & 0.147\scriptsize$\pm$.010 \\ 
       
        MEMOS (Ours)    &  0.87\scriptsize$\pm$.000 & \textbf{0.70\scriptsize$\mathbf{\pm}$.012} & 0.12\scriptsize$\pm$.037& \textbf{0.23\scriptsize$\mathbf{\pm}$.005} & 0.46\scriptsize$\pm$.172 & \textbf{0.65\scriptsize$\mathbf{\pm}$.045} & 0.35\scriptsize$\pm$.126\\ 
      \hline
        \end{tabular}
    \end{minipage}%
    \hfill
\end{table*}

We evaluate our method against the following baselines: directly thresholding the entropy of the base model's prediction (\textit{Entropy}), softmax thresholding (\textit{Softmax}) \cite{hendrycks2016baseline}, ensemble consensus over three networks \textit{(Ensemble)}, 
\textit{Learned Density}~\cite{blum2021fishyscapes} proposed by the authors of the Fishyscapes benchmark,  
ODIN~\cite{liang2017enhancing}, and Entropy Maximization using Coco dataset samples (\textit{EM-Coco})~\cite{chan2021entropy}. Note that we do not include generative baselines that require training additional large models, such as Synboost~\cite{di2021pixel}, as their inference time is orders of magnitude greater than our method and baselines, making them impractical for use with robots that require real-time perception loops. See Table \ref{tab:infer_time} for more details.

All approaches use the same base model, with the following modifications: \textit{Ensemble} trains three models over three random seeds, MaxEnt finetunes the base model on our synthetic OOD data, ODIN uses the base model with additional hyperparameters finetuned on an OOD validation dataset (reported in~\cite{liang2017enhancing}), and \textit{EM-Coco} uses its author-provided pre-trained weights. Additionally, we use the metric values reported by \cite{blum2021fishyscapes} for \textit{Learned Density}, since we were unable to find an implementation, and results were reported using the same base model as in this work. The metrics reported for all baselines are averaged over three random seeds\footnote{Learned Density and \textit{EM-Coco} and \textit{EM-Coco\&Meta} are reported over one random seed, as they use author-provided weights/results.}. All methods are evaluated on an input image size of 1024 x 2048.

We evaluate OOD detection performance using AUPRC, which gives greater importance to OOD detection by accounting for class imbalance. We treat OOD labels as the positive class. Additionally, we report the mean intersection-over-union (mIoU) for the ID cityscapes validation set, and the false positive rate for a 0.95 true positive rate (FPR-95) for OOD detection. These metrics serve as sanity checks to verify that the models can still perform effective semantic segmentation for ID classes while performing OOD detection.
We also report inference time in milliseconds.

\subsection{Results}

Over all of the benchmarks, our MEMOS framework outperforms all baselines that use no OOD data at training time, as shown in the top half of Table \ref{tab:results1}. As MEMOS is additive to the simple \textit{Entropy} baseline, significantly outperforming \textit{Entropy} indicates that our MaxEnt model and/or metacognitive network are beneficial for OOD detection, which we evaluate further in the ablation studies below. Additionally, when some OOD data \textit{is} available at training time, shown in the bottom half of Table \ref{tab:results1}, adding our metacognitive network to \textit{EM-Coco} improves performance over all other methods, showing both our modules's effectiveness for OOD detection and also validating its compatibility with other segmentation methods. Note that MEMOS without access to any OOD data also outperforms ODIN, and approaches the performance of \textit{EM-Coco} in the Lost and Found benchmark\footnote{For Fishyscapes benchmark, we attribute \textit{EM-Coco}'s high performance to data overlap with their additional OOD validation set, which contains classes similar to the OOD instances in the Fishyscapes datasets.}.


 ID detection on the Cityscapes validation, as shown by the mIoU column in Table \ref{tab:results1} demonstrates that our framework does not hinder ID semantic segmentation performance; there is no tradeoff when improving OOD detection performance. FPR-95 varies considerably across all methods and datasets, but when considered with the high mIoU performance on ID data, it shows that all of the methods we evaluated are able to perform OOD detection without sacrificing performance on ID data.

When considering open-world robotics applications~\cite{wong2020identifying, balasubramanian2021open, boteanu2015towards, feng2019challenges, kim2022learning, gunther2017toward, dey2023addressing}, short inference times are critical. We evaluate inference times for all methods on an NVIDIA RTX 3090. Table \ref{tab:infer_time} shows that many of the methods we evaluated, including our MEMOS framework, can be deployed in a real-time perception loop of 30-40 Hz. Notable exceptions are \textit{Ensemble} methods, which multiply the inference time by the ensemble size, and the significant computational burden of \textit{ODIN}, generative methods represented by \textit{Synboost}, and simply learning a larger end-to-end base network (\textit{Add'l Conv}, discussed further in the ablation studies below), which are all significantly less practical running closer to 1 Hz. Note, \textit{Add'l Conv}, was run on an NVIDIA RTX A6000 as the computational load was too high for a 3090.
 
\textbf{Ablation Studies:} We show contributions of MEMOS' individual components via ablation studies summarized in Table \ref{tab:ablation1}. Both the metacognitive network (\textit{Metacognitive-Only}) and MaxEnt components individually improve the performance of \textit{Entropy}. Combining both components improves performance further, showing that the two components are complementary. We take this as evidence that the entropy calibration of the MaxEnt model improves the detection ability of the metacognitive module, as our design intended.

We also create an additional baseline \textit{Add'l Conv.} by appending additional convolution layers to the base network, with as many parameters as the base plus metacognitive network (\textit{Metacognitive-Only}), and train it using the Cityscpaes dataset. While this baseline does marginally better than \textit{Entropy} due to its larger network size for the Lost and Found dataset, it performs worse than the comparably-sized \textit{Metacognitive-Only} across all datasets. This shows that the structure imposed by our metacognitive approach is beneficial beyond simply increasing the number of network parameters. Further, designing a metacognitive module as a separate network component reduces computational burden as shown in Table \ref{tab:infer_time}.

\section{LIMITATIONS}
\label{sec:limitations}

Our framework assumes that training on synthetic OOD data will generalize sufficiently to actual OOD samples. Our main failure mode is sensitivity to well-calibrated networks---any base model with a poorly calibrated entropy will likely limit the metacognitive network's efficacy. 
We examined only Gaussian blurring to generate the synthetic OOD images, and suggest evaluating additional image transformations in future work. While we demonstrate significant increase in performance over comparable baselines on standard benchmarks, this work would benefit from real-world OOD detection experiments situated on physical robots to overcome the limitations of such constructed benchmarks.

\section{CONCLUSIONS}
\label{sec:conclusion}
We presented the MEMOS framework, consisting of a novel metacognitive network module that can improve OOD detection for state-of-the-art semantic segmentation models by leveraging uncertainty measures and spatial information. We also demonstrate that the performance of the metacognitive module improves considerably when fine-tuning the base segmentation model using maximum entropy training over synthetic OOD data generated in context with ID data, outperforming state-of-the-art OOD detection for segmentation baselines trained with equivalent access to ID data and realistic restrictions on available OOD data. Finally, we show that our framework has a low inference time that is suitable for real-time perception.


\balance








\bibliographystyle{IEEEtran}
\bibliography{IEEEabrv,ICRA2024/references}

\end{document}